\pdfoutput=1
\documentclass[11pt]{article}

\usepackage[final]{EMNLP2022}

\usepackage{times}
\usepackage{latexsym}

\usepackage[T1]{fontenc}
\usepackage[utf8]{inputenc}

\usepackage{microtype}

\usepackage{inconsolata}

\usepackage{amsmath,amsfonts,bm}

\def\eqref#1{equation~\ref{#1}}
\def\1{\bm{1}}

\DeclareMathAlphabet{\mathsfit}{\encodingdefault}{\sfdefault}{m}{sl}
\SetMathAlphabet{\mathsfit}{bold}{\encodingdefault}{\sfdefault}{bx}{n}

\usepackage{hyperref}
\usepackage{url}

\usepackage{multirow}
\usepackage{makecell}
\usepackage{graphicx}
\usepackage{tablefootnote}
\usepackage{adjustbox}
\usepackage{subfiles}
\usepackage{booktabs}

\usepackage{wrapfig}
\usepackage{subfigure}
\usepackage{latexsym}

\definecolor{limegreen}{RGB}{76,153,0}

\usepackage{CJKutf8}

\title{Enhancing Multilingual Language Model with \\ Massive Multilingual Knowledge Triples}

\author{
Linlin Liu$^{1,2}$\thanks{\; Linlin Liu is under the Joint PhD Program between Alibaba and Nanyang Technological University.} ~~ 
Xin Li$^1$ ~
Ruidan He$^1$ ~
Lidong Bing$^1$\thanks{\; Corresponding author.} ~
Shafiq Joty$^{2,3}$ ~
Luo Si$^1$\\
$^1$DAMO Academy, Alibaba Group\\
$^2$Nanyang Technological University, Singapore\\
$^3$Salesforce Research\\
$^1$\{linlin.liu, xinting.lx, ruidan.he, l.bing, luo.si\}@alibaba-inc.com 
$^2$srjoty@ntu.edu.sg
}

\begin{document}
\maketitle

\begin{abstract}

Knowledge-enhanced language representation learning has shown promising results across various knowledge-intensive NLP tasks. However, prior methods are limited in efficient utilization of multilingual knowledge graph (KG) data for language model (LM) pretraining. They often train LMs with KGs in indirect ways, relying on extra entity/relation embeddings to facilitate knowledge injection. In this work, we explore methods to make better use of the multilingual annotation and language agnostic property of KG triples, and present novel knowledge based multilingual language models (KMLMs) trained directly on the knowledge triples. We first generate a large amount of multilingual synthetic sentences using the Wikidata KG triples. Then based on the intra- and inter-sentence structures of the generated data, we design pretraining tasks to enable the LMs to not only memorize the factual knowledge but also learn useful logical patterns. Our pretrained KMLMs demonstrate significant performance improvements on a wide range of knowledge-intensive cross-lingual tasks, including named entity recognition (NER), factual knowledge retrieval, relation classification, and a newly designed logical reasoning task.\footnote{Our code, data and pretrained models are available at \url{https://github.com/ntunlp/kmlm.git}.}

\end{abstract}

\section{Introduction}

Pretrained Language Models (PLMs) such as BERT~\citep{devlin-etal-2019-bert} and RoBERTa~\citep{liu2019roberta} have achieved superior performances on a wide range of NLP tasks. Existing PLMs usually learn universal language representations from general-purpose large-scale corpora but do not concentrate on capturing world's factual knowledge. It has been shown that knowledge graphs (KGs), such as Wikidata~\citep{wikidata14} and Freebase \citep{bollacker2008freebase}, can provide rich factual information for better language understanding. Many studies have demonstrated the effectiveness of incorporating such factual knowledge into monolingual PLMs \citep{peters-etal-2019-knowledge,zhang-etal-2019-ernie,liu2019kbert,poerner-etal-2020-e,wang-etal-2021-k}. Following this, a few recent attempts have been made to enhance multilingual PLMs with Wikipedia or KG triples \citep{calixto-etal-2021-wikipedia,ri-etal-2022-mluke,jiang2021xlm}.
However, due to the structural difference between KG and texts, existing KG based pretraining often relies on extra relation/entity embeddings or additional KG encoders for knowledge enhancement. These extra embeddings/components may add significantly more parameters which in turn increase inference complexity, or cause inconsistency between pre-train and downstream tasks. For example, mLUKE \citep{ri-etal-2022-mluke} has to enumerate all possible entity spans for NER to minimize the inconsistency caused by entity and entity position embeddings. Other methods \citep{liu2019kbert,jiang2021xlm} also require KG triples to be combined with relevant natural sentences as model input during training or inference.

In this work, we propose KMLM, a novel \textbf{K}nowledge-based \textbf{M}ultilingual \textbf{L}anguage \textbf{M}odel pretrained on massive multilingual KG triples. Unlike prior knowledge enhanced models~\citep{zhang-etal-2019-ernie,peters-etal-2019-knowledge,liu2019kbert,wang-etal-2021-k}, our model requires neither a separate encoder to encode entities/relations, nor heterogeneous information fusion to fuse multiple types of embeddings (e.g., entities from KGs and words from sentences).
The key idea of our method is to convert the structured knowledge from KGs to sequential data which can be directly fed as input to the LM during pretraining. 
Specifically, we generate three types of training data -- the \emph{parallel knowledge data}, the \emph{code-switched knowledge data} and the \emph{reasoning-based data}. The first two are obtained by generating parallel or code-switched sentences from triples of Wikidata ~\citep{wikidata14}, a collaboratively edited multilingual KG. 
The reasoning-based data, containing rich logical patterns, is constructed by converting cycles from Wikidata into word sequences in different languages. We then design pretraining tasks that are operated on the parallel/code-switched data to memorize the factual knowledge across languages, and on the reasoning-based data to learn the logical patterns.

Compared to existing knowledge-enhanced pretraining methods~\citep{zhang-etal-2019-ernie,liu2019kbert,peters-etal-2019-knowledge,jiang2021xlm}, KMLM has the following key advantages. (1) KMLM is explicitly trained to derive new knowledge through logical reasoning. Therefore, in addition to memorizing knowledge facts, it also learns the logical patterns from the data. (2) KMLM does not require a separate encoder for KG encoding, and eliminates relation/entity embeddings, which enables KMLM to be trained on a larger set of entities and relations without adding extra parameters. The token embeddings are enhanced directly with knowledge related training data. (3) KMLM does not rely on any entity linker to link the text to the corresponding KG entities, as done in existing methods \citep{zhang-etal-2019-ernie,peters-etal-2019-knowledge,poerner-etal-2020-e}. This ensures KMLM to utilize more KG triples even if they are not linked to any text data, and avoids noise caused by incorrect links. (4) KMLM keeps the model structure of the multilingual PLM without introducing any additional component during both training and inference stages. This makes the training  much easier, and the trained model is directly applicable to downstream NLP tasks. 

We evaluate KMLM on a wide range of knowledge-intensive cross-lingual tasks, including NER, factual knowledge retrieval, relation classification, and logical reasoning which is a novel task designed by us to test the reasoning capability of the models. 
Our KMLM achieves consistent and significant improvements on all knowledge-intensive tasks, meanwhile it does not sacrifice the performance on general NLP tasks.

\section{Related Work}

Knowledge-enhanced language modeling aims to incorporate knowledge, concepts and relations into the PLMs~\citep{devlin-etal-2019-bert,liu2019roberta,brown2020language}, which proved to be beneficial to language understanding~\citep{talmor-etal-2020-olmpics}. 

The existing approaches mainly focus on monolingual PLMs, which can be roughly divided into two lines: implicit knowledge modeling and explicit knowledge injection. Previous attempts on implicit knowledge modeling usually consist of entity-level masked language modeling~\citep{sun2019ernie,liu2019kbert}, entity-based replacement prediction~\citep{xiong2020pretrained}, knowledge embedding loss as regularization~\citep{10.1162/tacl_a_00360} and universal knowledge-text prediction~\citep{DBLP:journals/corr/abs-2107-02137}. In contrast to implicit knowledge modeling, the methods of explicit knowledge injection separately maintain a group of parameters for representing structural knowledge. Such methods \citep{zhang-etal-2019-ernie} usually require a heterogeneous information fusion component to fuse multiple types of embeddings obtained from the text and KGs. \citet{zhang-etal-2019-ernie} and \citet{poerner-etal-2020-e} employ external entity linker to discover the entities in the text and perform feature interaction between the token embeddings and entity embeddings during the encoding phase of a transformer model. \citet{peters-etal-2019-knowledge} borrow the pre-computed knowledge embeddings as the supporting features of training an internal entity linker. \citet{wang-etal-2021-k} insert an adapter component in each transformer layer to store the learned factual knowledge. 

Extending knowledge based pretraining methods to the multilingual setting has received increasing interest recently. \citet{zhou-etal-2022-prix} propose an auto-regressive model trained on knowledge triples for multilingual KG completion. \citet{calixto-etal-2021-wikipedia,ri-etal-2022-mluke} attempt improving multilingual entity representation via Wikipedia hyperlink prediction, however, their methods add a large amount of parameters due to the reliance on extra entity embeddings. For example, the mLUKE$_\textrm{BASE}$ \citep{ri-etal-2022-mluke} model initialized with XLM-R$_\textrm{BASE}$ doubles the number of parameters (586M vs 270M). Similar to us, \citet{jiang2021xlm} also utilize KG for PLM pretraining. They employ KG and Wikipedia entity descriptions to inject knowledge into multilingual LM, but relation embeddings are also required to assist learning.

Moreover, the above methods only focus on memorizing the existing facts but ignore the reasoning over the unseen/implicit knowledge that is derivable from the existing facts. Such reasoning capability is regarded as a crucial part of building consistent and controllable knowledge-based models~\citep{NEURIPS2020_e992111e}. In this paper, our explored methods for multilingual knowledge-enhanced pretraining boost the capability of implicit knowledge reasoning, together with the purpose of consolidating knowledge modeling and multilingual pretraining \citep{mulcaire-etal-2019-polyglot,conneau-etal-2020-unsupervised}.

\section{Framework}

In this section, we describe the proposed framework for knowledge based multilingual language model (KMLM) pretraining. We first describe the process to generate knowledge-intensive multilingual training data, followed by the pretraining tasks to train the language models to memorize factual knowledge and learn logical patterns from the generated data.

\subsection{Knowledge Intensive Training Data}
\label{sec:knowledge_intensive_data}

\begin{table}[t!]
    \centering
    \scalebox{0.75}{
    \begin{tabular}{llll}
    \toprule
    %\Xhline{3\arrayrulewidth}
        \textbf{ID} & \textbf{Language} & \textbf{Label} & \textbf{Aliases}  \\ \midrule
        \multirow{4}{*}{Q1420} & English & motor car & auto, autocar, \dots \\
        & Spanish & automóvil & coche, carro, \dots \\
        & Hungarian & autó & gépkocsi, személyautó, \dots \\
        & \dots & \dots & \dots \\
    %\Xhline{3\arrayrulewidth}
    \bottomrule
    \end{tabular}}
    \caption{An example (\url{https://www.wikidata.org/wiki/Q1420}) of the Wikidata entity labels and aliases in multiple languages. Q1420 is the unique entity ID.}
    \label{tab:framework_wikidata}
\end{table}

In addition to the large-scale plain text corpus that is commonly used for language model pretraining, we also generate a large amount of knowledge intensive training data from Wikidata \citep{wikidata14}, a publicly accessible knowledge base edited collaboratively. Wikidata is composed of massive amounts of KG triples $(h,r,t)$, where $h$ and $t$ are the head and tail entities respectively, $r$ is the relation type. As shown in Table~\ref{tab:framework_wikidata}, most of the entities, as well as the relations in Wikidata, are annotated in multiple languages. In each language, many aliases are also given though some of them are used infrequently.

\paragraph{Code-Switched Synthetic Sentences}
Training language models on high-quality code-switched sentences is one of the most intuitive ways to learn language agnostic representation \citep{winata-etal-2019-code}, where the translations of words/phrases can be treated in a similar way as their aliases. The code mixing techniques have also proved to be helpful for improving cross-lingual transfer performance in many NLP tasks \citep{qin2020cosdaml,santy2021bertologicomix}. Therefore, we propose a novel method to generate code-switched synthetic sentences using the multilingual KG triples. See Fig.~\ref{fig:framework_code_switched} for some generated examples. 

\begin{figure}[t!]
    \centering
    \resizebox{0.95\linewidth}{!}{    
    \adjustbox{minipage=[r][17em][b]{0.6\textwidth},scale={0.95}}{
    \textcolor{blue}{Original (en):} \\
    (motor car, designed to carry, passenger) \\[-0.5em]
    
    \textcolor{blue}{Code-Switched (en-fr):} \\
    motor car [mask] \textbf{conçu pour transporter} [mask] passenger. \\
    motor car [mask] designed to carry [mask] \textbf{passager}. \\[-0.5em]
    
    \textcolor{blue}{Code-Switched (en-fr) \& Alias-Replaced:} \\
    \underline{automobile} [mask] \underline{\textbf{destiné au transport}} [mask] passenger. \\
    motor car [mask] \underline{intended to carry} [mask] \textbf{passager}. \\ [-0.5em]
    
    \textcolor{blue}{Parallel (en-fr) \& Alias-Replaced:} \\
    \underline{autocar} [mask] designed to carry [mask] passenger. \\
    \underline{\textbf{voiture}} [mask] \textbf{conçu pour transporter} [mask] \textbf{passager}. \\
    }}
    \caption{Examples of the en-fr code-switched and parallel synthetic sentences. The words replaced with translations or aliases are marked with bold font and underline, respectively.}
    \label{fig:framework_code_switched}
\end{figure}

For each triple $(h,r,t)$ in Wikidata, we use $h_{l,0}$ to denote the default label of $h$ in language $l$. 
For the entity Q1420 in Table~\ref{tab:framework_wikidata}, $h_{en,0}$ is ``motor car'' and $h_{es,0}$ is ``automóvi''. $h_{l,i}$ denotes the aliases when the integer $i>0$.
We define $r_{l,i}$ and $t_{l,i}$ in the same way for the relation and the tail entity, respectively. 
Since English is resource-rich and often treated as the source language for cross-lingual transfer, we only consider language pairs of $\{(en, l^\prime)\}$ for code switching, where $l^\prime$ is an arbitrary non-English language. With such a design, English can also work as a bridge for cross-lingual transfer between a pair of none English languages. 

Specifically, the code-switched sentences for $(h,r,t)$ can be generated in 4 steps: 1) select a language pair $(en, l^\prime)$; 2) find the English default labels $(h_{en,0},r_{en,0},t_{en,0})$; 3) For each item in the triple, uniformly sample a value $v \in \{true, false\}$, if $v$ is $true$ and the item has a translation (i.e. default label) in $l^\prime$, then replace the item with the translation in $l^\prime$; 4) generate the sequence of \textit{``h [mask] r [mask] t.''} by inserting two mask tokens. The alias-replaced sentences can be generated in a similar way, except that we randomly sample aliases in the desired language to replace the default label in steps 2 and 3.

\paragraph{Parallel Synthetic Sentences}
Parallel data has also been widely exploited to improve cross-lingual transfer \citep{arivazhagan2019massively,conneau2019cross,chi-etal-2021-infoxlm}. However, it is expensive to obtain a large amount of parallel data for LM pretraining. We propose a method to generate a large amount of knowledge intensive parallel synthetic sentences, with a minor modification of the method for generating code-switched sentences described above. For each triple $(h,r,t)$ extracted from Wikidata, the corresponding synthetic sentences in different languages can be generated by first finding the default labels $(h_{l,0},r_{l,0},t_{l,0})$ for each language $l$, and then inserting mask tokens to generate sequences in the form \textit{``h [mask] r [mask] t.''}. Fig.~\ref{fig:framework_code_switched} shows an example. More sentences can be generated by replacing the default labels with their aliases.

%\begin{wrapfigure}{r}{0.6\textwidth}
\begin{figure}[t!]
\centering
\subfigure[Cycle of length 3.
\label{fig:framework_cycles3}]{\includegraphics[scale=0.38]{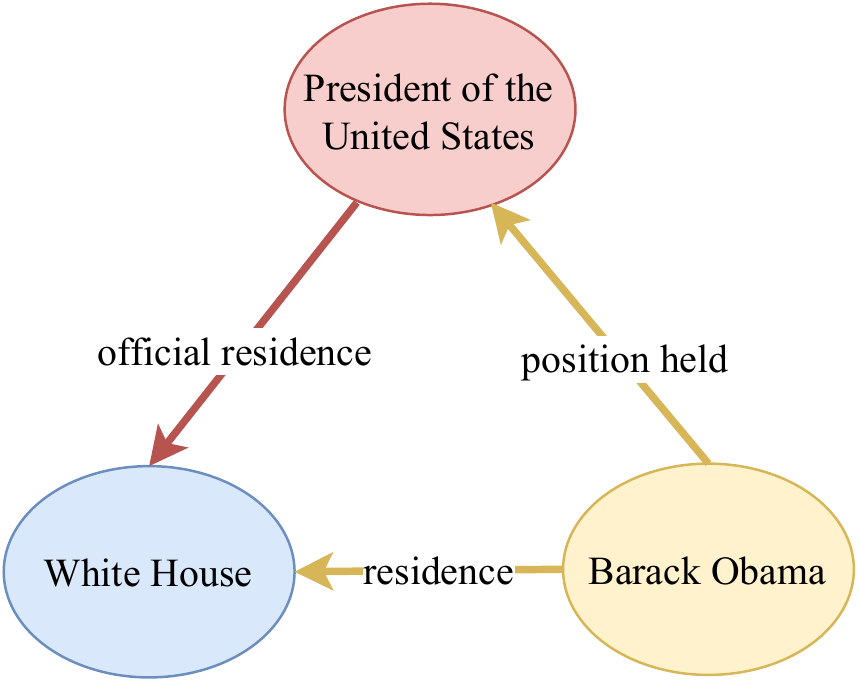}}
\hspace{-0.3em}
\subfigure[Cycle of length 4.
\label{fig:framework_cycles4}]{\includegraphics[scale=0.38]{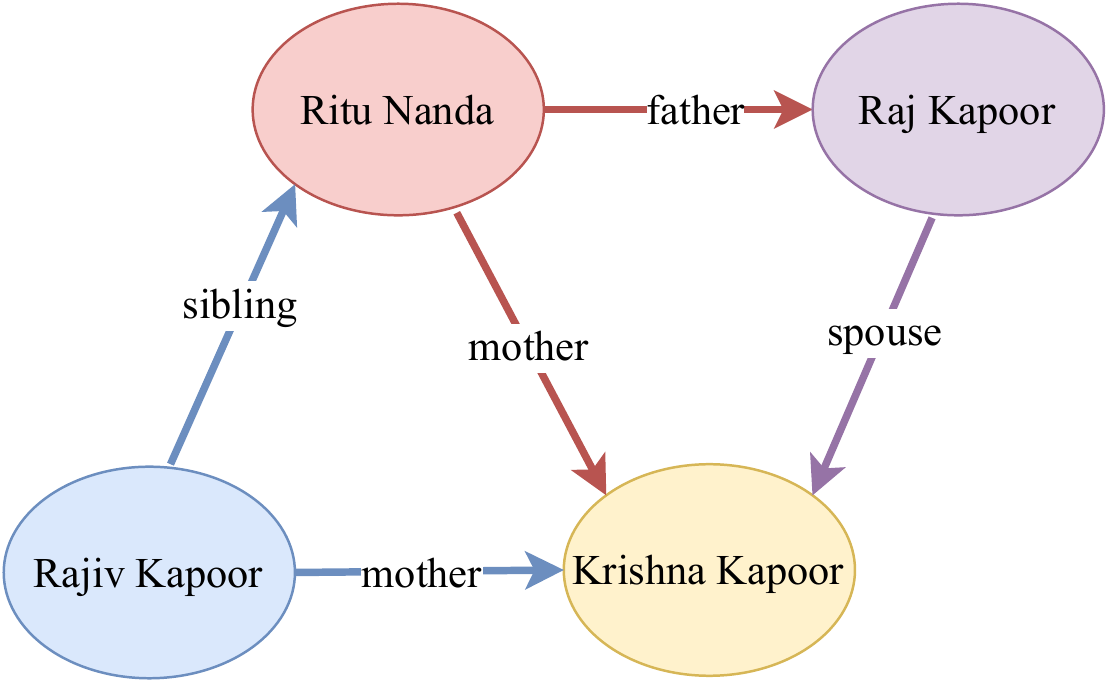}}
\caption{ Examples of extracted cycles of length 3 and 4.}
\label{fig:framework_cycles}
\end{figure}
%\end{wrapfigure}

\paragraph{Reasoning-Based Training Data}

The capability of logical reasoning allows humans to solve complex problems with limited information. However, this ability did not receive much attention in the previous LM pretraining methods. In KGs, we can use nodes to represent entities, and edges between any two nodes to represent their relations.  In order to train the model to learn logical patterns, we generate a large amount of reasoning-based training data by finding cycles from the Wikidata KG. As shown with an example in Fig.~\ref{fig:framework_cycles3}, the cycles of length 3 can be viewed as the basic component for more complex logical reasoning process. We train language models to learn the entity-relation co-occurrence patterns so as to infer the best candidate relations for incomplete cycles, i.e. deriving the implicit information from the given context. 

Similar to the structure of the parallel/code-switched synthetic sentences described above, the cycles in Fig.~\ref{fig:framework_cycles3} is composed of 3 triples, and hence can be converted to 3 synthetic sentences (the first example in Fig.~\ref{fig:framework_task_logic}). To increase the difficulty, we also extract cycles of length 4 to generate the reasoning oriented training data. However, we find that simply increasing the length of cycles makes the samples less logically coherent. Thus, we add an extra constraint that each length-4 cycle is required to have at least one additional diagonal edge. Fig.~\ref{fig:framework_cycles4} shows such an example. It can be converted to a training sample of 5 sentences in the same way as above. For the multilingual reasoning-based data, we only generate monolingual sentences, i.e. without applying code mixing.

We treat Wikidata as an undirected graph when extracting cycles. Given an entity, the length-3 cycles containing this entity can be easily extracted by first finding all the neighbouring entities, and then iterating through the pairs of neighbouring entities to check whether they are also connected. The length-4 cycles with an additional diagonal edge connecting any two neighbours can be extracted with a few extra steps. Assuming we have identified a length-3 cycle containing entity $A$ and its two neighbouring entities $B$ and $C$, we can iterate through the neighbours of $B$ (excluding $A$ and $C$) to check whether it is also connected to $C$. We remove the duplicate cycles in data generation.

\subsection{Pretraining Tasks}
\label{sec:pretrain_task}

%\begin{wrapfigure}{r}{0.55\textwidth}
\begin{figure}[t!]
\centering
\includegraphics[scale=0.58]{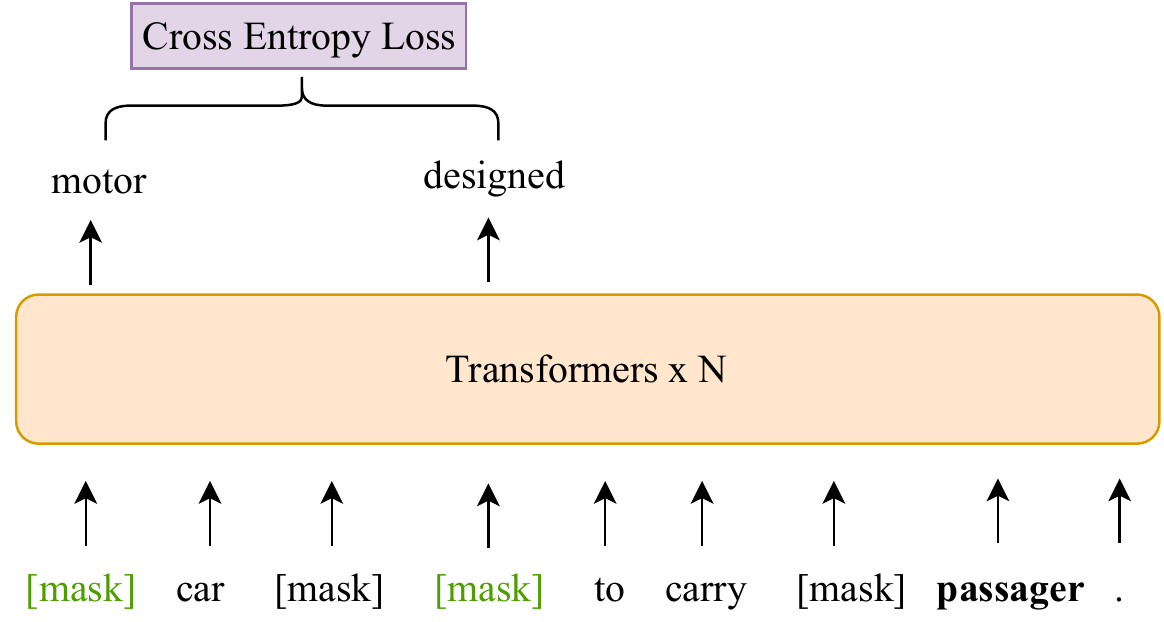}

\caption{MLM on the code-switched synthetic sentence \textit{``motor car [mask] designed to carry [mask] \textbf{passager}.''}. The cross entropy loss $\mathcal{L}_{K}$ for \textbf{K}nowledge Oriented Pretraining is only computed over the randomly masked entity and relation tokens highlighted in lime green. For simplicity, the sub-word tokens are not shown in this example.}
\label{fig:framework_task_mlm}
\end{figure}

\paragraph{Multilingual Knowledge Oriented Pretraining}
In the generated code-switched and parallel synthetic sentences, the \textit{``[mask]''} tokens are added between entities and relations to denote the linking words. For example, the first mask token in \textit{``motor car [mask] designed to carry [mask] \textbf{passager}.''} may denote \textit{``is''}, while the second one may denote \textit{``certains''} (French word ``certains'' means ``some'' or ``certain''). Since the ground truth of such masked linking words are not known, we do not compute the loss for those corresponding predictions. Instead, we randomly mask the remaining tokens in the parallel/code-switched synthetic sentence, and compute the cross entropy loss over these masked entity and relation tokens (Fig.~\ref{fig:framework_task_mlm}). We use $\mathcal{L}_{K}$ to denote this cross entropy loss for \textbf{K}nowledge Oriented Pretraining. Note that our models are not trained on the sentence pairs like the Translation LM loss or TLM \citep{conneau2019cross} when utilizing the parallel or code-switched pairs. Alternatively, we shuffle the data, and feed one sentence into the model each time (as shown in Fig.~\ref{fig:framework_task_mlm}), which makes our model inputs more consistent with those of the downstream tasks.

\begin{figure*}[ht!]
    \centering
    \resizebox{0.9\linewidth}{!}{    
    \adjustbox{minipage=[r][16em][b]{1.4\textwidth},scale={0.9}}{
    \textbf{Length-3 Cycle:} \\
    \textcolor{blue}{Original:}
    President of the United States [mask] \textcolor{orange}{official residence} [mask] White House. Barack Obama [mask] \textcolor{limegreen}{residence} [mask] White House.  Barack Obama [mask] \textcolor{orange}{position held} [mask] President of the United States. \\
    
    \textcolor{blue}{Masked:}
    President of the United States [mask] \textcolor{orange}{official residence} [mask] White House. Barack Obama [mask] \textcolor{limegreen}{[mask]} [mask] White House.  Barack Obama [mask] \textcolor{orange}{position held} [mask] President of the United States.\\
    
    \textbf{Length-4 Cycle:} \\
    \textcolor{blue}{Original:}
    Ritu Nanda [mask] \textcolor{orange}{father} [mask] Raj Kapoor. Ritu Nanda [mask] \textcolor{orange}{mother} [mask] \textcolor{limegreen}{ Krishna Kapoor}. Rajiv Kapoor [mask] \textcolor{orange}{mother} [mask] Krishna Kapoor. Rajiv Kapoor [mask] \textcolor{limegreen}{sibling} [mask] Ritu Nanda. Raj Kapoor [mask] \textcolor{orange}{spouse} [mask] Krishna Kapoor.\\
    
    \textcolor{blue}{Masked:}
    Ritu Nanda [mask] \textcolor{orange}{father} [mask] Raj Kapoor. Ritu Nanda [mask] \textcolor{orange}{mother} [mask] \textcolor{limegreen}{ [mask] } \textcolor{limegreen}{[mask]}. Rajiv Kapoor [mask] \textcolor{orange}{mother} [mask] Krishna Kapoor. Rajiv Kapoor [mask] \textcolor{limegreen}{[mask]} [mask] Ritu Nanda. Raj Kapoor [mask] \textcolor{orange}{spouse} [mask] Krishna Kapoor. \\
    }}
    \caption{Examples of the masked training samples for logical reasoning. The relations are highlighted in orange. The masked entity and relation tokens are highlighted in lime green.}
    \label{fig:framework_task_logic}
\end{figure*}

\paragraph{Logical Reasoning Oriented Pretraining}

We design tasks to train the model to learn logical reasoning patterns from the synthetic sentences generated from the length-3 and length-4 cycles. As can be seen in Fig.~\ref{fig:framework_task_logic}, both of the relation prediction and entity prediction problems are cast as masked language modeling. For the length-3 cycles, each entity appears exactly twice in every training sample. Formulating the task as a masked entity prediction problem may lead to shortcut learning \citep{geirhos2020shortcut} by simply counting the appearance numbers of the entities. Therefore, we only mask one random relation in each sample for model training, and let the model learn to predict the masked relation tokens based on the context.

Two types of tasks are designed to train the model to learn reasoning with the length-4 cycles: 1) For 80\% of the time, we train the model to predict randomly masked relation and entities. We first mask one random relation. To increase the difficulty, we also mask one or two randomly selected entities at equal chance. 
The lower half of Fig.~\ref{fig:framework_task_logic} shows an example where one relation and one entity are masked. 2) For the remaining 20\% of the time, we randomly mask a whole sentence to let the model learn to derive new knowledge from the remaining context. To provide some hints on the expected new knowledge, we keep the relation of the selected sentence unmasked, i.e., only mask its two entities. The loss $\mathcal{L}_{L}$ for \textbf{L}ogical Reasoning Oriented Pretraining can also be computed with the cross entropy loss over the masked tokens. Note that masked entity prediction is not always non-trivial in this task. For example, when we mask exactly one entity and the entity $E$ only appears once in the masked sample, then it is easy to guess $E$ is the masked one. In Fig.~\ref{fig:framework_task_logic}, a concrete example is masking the first appearance of \textit{``Raj Kapoor''} in the original sentence of the length-4 cycle. We do not deliberately avoid such cases, since they may help introduce more diversity to the training data.

\paragraph{Loss Function}
In addition to the pretraining tasks designed above, we also train the model on the plain text data with the original masked language modeling loss $\mathcal{L}_{MLM}$ used in previous work \citep{devlin-etal-2019-bert,conneau-etal-2020-unsupervised}. Therefore, the final loss can be computed as:
\begin{equation}
\mathcal{L} = \mathcal{L}_{MLM} + \alpha (\mathcal{L}_{K} +  \mathcal{L}_{L})
\label{eq:loss_final}
\end{equation}
where $\alpha$ is a hyper-parameter to adjust the weights of the original MLM and the losses for modeling the multilingual knowledge and logical reasoning.

\section{Experiments}
We first describe the pretraining details of our KMLMs. Then we verify its effectiveness on the knowledge-intensive tasks. Finally, we examine its performance on general cross-lingual tasks. In all of the tasks except X-FACTR~\citep{jiang-etal-2020-x}, the PLMs are fine-tuned  on the English training set and then evaluated on the target language test sets. The evaluation results are averaged over 3 runs with different random seeds. X-FACTR does not require fine-tuning, so the PLMs are directly evaluated using the official code. The results of the baseline models are reproduced in the same environment.

\subsection{Pretraining Details} \label{subsec:pre}
Our proposed framework can be conveniently implemented on top of the existing transformer encoder based models like mBERT \citep{devlin-etal-2019-bert} and XLM-R \citep{conneau-etal-2020-unsupervised} without any modification to the model structure. Therefore, instead of pretraining the model from scratch, it is more time- and cost-efficient to initialize the model with the checkpoints of existing pretrained models. We build our knowledge intensive training data in 10 languages: English, Vietnamese, Dutch, German, French, Italian, Spanish, Japanese, Korean and Chinese. We only use the 5M entities and 822 relations filtered by \citet{10.1162/tacl_a_00360}, and generate 250M code-switched synthetic sentences, 190M parallel synthetic sentences\footnote{Half of the parallel and code-switched sentences are also alias-replaced. The size of the generated parallel data is smaller than the code-switched one because some of the entities/relations do not have 
annotation in the target language.} and 100M reasoning-based samples following the steps in \S\ref{sec:knowledge_intensive_data}. In addition, 260M sentences are sampled from the CC100 corpus\footnote{\url{http://data.statmt.org/cc-100/}} \citep{wenzek-etal-2020-ccnet} for the 10 languages. Our models KMLM-XLM-R$_\textrm{BASE}$ and KMLM-XLM-R$_\textrm{LARGE}$ are initialized with XLM-R$_\textrm{BASE}$ and XLM-R$_\textrm{LARGE}$, respectively. Then we continue to pretrain these models with the proposed tasks (\S\ref{sec:pretrain_task}). KMLM$_\textrm{CS}$, KMLM$_\textrm{Parallel}$ and KMLM$_\textrm{Mix}$ are used to differentiate the models trained on the code-switched data, parallel data and the concatenated data of these two, respectively. The reasoning-based data is used in all these three models, and ablation studies are presented in \S\ref{sec:crosslingual_logical_reason} to verify the effectiveness of logical reasoning task.

Previous studies showed that the original mBERT model outperforms XLM-R on the X-FACTR~\citep{jiang-etal-2020-x} and RELX~\citep{koksal-ozgur-2020-relx} tasks, so we also initialize KMLM-mBERT$_\textrm{BASE}$ with mBERT$_\textrm{BASE}$, and train it on Wikipedia corpus for a more faithful comparison\footnote{The original mBERT$_\textrm{BASE}$ is trained using the Wikipedia.}. We find the KMLM$_\textrm{CS}$ and KMLM$_\textrm{Mix}$ models initialized with the XLM-R$_\textrm{BASE}$ checkpoint outperform the corresponding KMLM$_\textrm{Parallel}$ model in most of the tasks, so we only train KMLM$_\textrm{CS}$ and KMLM$_\textrm{Mix}$ when comparing with XLM-R$_\textrm{LARGE}$ and mBERT$_\textrm{BASE}$. See Appendix \S\ref{sec:app_pretrain} for more pretraining details.

\subsection{Cross-lingual Named Entity Recognition}
\label{exp:ner}

\begin{table}[t!]
    \centering
    \scalebox{0.56}{
    \begin{tabular}{lcccccc}
    \Xhline{3\arrayrulewidth}
        & en & de & nl & es & avg$_{tgt}$ & $\Delta_{avg}$  \\ 
        \hline
        mBERT$_\textrm{BASE}$${^\dag}$  & 90.6 & 69.2 & 77.9 & 75.4 & 74.2 & - \\
        XLM-K${^\ddag}$ & 90.7 & 72.9 & 80.3 & 75.2 & 76.1 & - \\
        XLM-R$_\textrm{BASE}$ & 91.16 & 68.87 & 79.00 & \textbf{76.70} & 74.86 & 0 \\ 
        KMLM$_\textrm{CS}$-XLM-R$_\textrm{BASE}$ (ours) & \textbf{91.47} & 73.52 & 80.95 & 76.59 & 77.02 & +2.16  \\
        KMLM$_\textrm{Parallel}$-XLM-R$_\textrm{BASE}$ (ours) & 91.44 & \textbf{73.96} & 81.06 & 75.94 & 76.99 & +2.13 \\
        KMLM$_\textrm{Mix}$-XLM-R$_\textrm{BASE}$ (ours) & 91.38 & 73.95 & \textbf{81.29} & 76.17 & \textbf{77.14} & +2.28 \\
        \hline
        XLM-R$_\textrm{LARGE}$ & 92.98 & 73.79 & 82.00 & 79.33 & 78.37 & 0 \\
        KMLM$_\textrm{CS}$-XLM-R$_\textrm{LARGE}$ (ours) & 92.81 & 76.22 & \textbf{84.12} & 78.63 & 79.66 & +1.29 \\
        KMLM$_\textrm{Mix}$-XLM-R$_\textrm{LARGE}$ (ours) & \textbf{93.12} & \textbf{76.88} & 82.84 & \textbf{80.08} & \textbf{79.93} & +1.56\\
    \Xhline{3\arrayrulewidth}
    \end{tabular}}
    \caption{Zero-shot cross-lingual NER F1 on the CoNLL02/03 datasets. The average results of non-English languages are reported in column avg$_{tgt}$. ${^\dag}$ The results are from \citep{liang-etal-2020-xglue}. ${^\ddag}$ The results are from \citep{jiang2021xlm}.}
    \label{tab:exp_ner_conll}
\end{table}

\begin{table*}[ht!]
    \centering
    \scalebox{0.76}{
    \begin{tabular}{lcccccccccccc}
    \Xhline{3\arrayrulewidth}
        & en & vi & nl & de & fr & it & es & ja & ko & zh & avg$_{tgt}$ & $\Delta_{avg}$ \\ 
        \hline
        
        \hline
        XLM-K & 83.32 & \textbf{72.80} & 81.39 & 76.96 & 78.75 & 78.81 & 71.60 & 17.14 & \textbf{57.75} & 19.68 & 61.65 & - \\
        XLM-R$_\textrm{BASE}$ & 82.59 & 68.09 & 80.08 & 74.71 & 76.50 & 77.06 & 71.05 & 20.34 & 48.46 & \textbf{26.32} & 60.29 & 0 \\
        KMLM$_\textrm{CS}$-XLM-R$_\textrm{BASE}$ (ours) & 83.43 & 70.55 & 82.18 & \textbf{77.87} & 79.19 & \textbf{80.06} & 75.96 & 19.32 & 57.54 & 20.95  & 62.62 & +2.33 \\
        KMLM$_\textrm{Parallel}$-XLM-R$_\textrm{BASE}$ (ours) & \textbf{83.54} & 70.93 & \textbf{82.30} & 77.79 & 78.40 & 79.83 & 76.06 & 18.16 & 57.40 & 20.44 & 62.37 & +2.08 \\
        KMLM$_\textrm{Mix}$-XLM-R$_\textrm{BASE}$ (ours) & 83.42 & 70.24 & 82.22 & 77.30 & \textbf{79.93} & 80.03 & \textbf{76.72} & \textbf{20.78} & 56.70 & 22.49 & \textbf{62.93} & +2.64 \\
        \hline
        XLM-R$_\textrm{LARGE}$ & 84.34 & 77.61 & 83.72 & 78.92 & 79.93 & 81.24 & 73.59 & 18.94 & 59.27 & \textbf{28.35} & 64.62 & 0 \\ 
        KMLM$_\textrm{CS}$-XLM-R$_\textrm{LARGE}$ (ours) & \textbf{85.07} & 77.89 & 84.55 & \textbf{81.32} & \textbf{83.65} & \textbf{82.57} & \textbf{78.93} & 14.95 & 60.68 & 19.43 & 64.89 & +0.27 \\
        KMLM$_\textrm{Mix}$-XLM-R$_\textrm{LARGE}$ (ours) & 84.87 & \textbf{77.99} & \textbf{84.62} & 81.13 & 82.85 & 82.28 & 77.30 & \textbf{21.22} & \textbf{61.88} & 26.69 & \textbf{66.22} & +1.60 \\ 
    \Xhline{3\arrayrulewidth}
    \end{tabular}
    }
    \caption{Zero-shot cross-lingual NER F1 on the WikiAnn dataset.}
    \label{tab:exp_ner_wikiann}
\end{table*}

Named entity recognition (NER) involves identifying and classifying named entities from unstructured text data. The elimination of entity/relation embeddings allows our models to be trained directly on a larger amount of entities without adding extra parameters or increasing computation cost. Direct training on entity-intensive synthetic sentences may also help improving entity representation more efficiently. We conduct experiments on the CoNLL02/03 \citep{tjong-kim-sang-2002-introduction,tjong-kim-sang-de-meulder-2003-introduction} and WikiAnn \citep{pan-etal-2017-cross} NER data to verify the effectiveness of our framework. The same transformer-based NER model and hyper-parameters as \citet{hu2020xtreme} are used in our experiments.

The results on CoNLL02/03 data are presented in Table~\ref{tab:exp_ner_conll}. Compared with XLM-R$_\textrm{BASE}$, all of our corresponding models improve the average F1 on target languages by more than 2.13 points. Especially on German, all of our models demonstrate at least 4.65 absolute gains. Moreover, all of our models also outperform XLM-K \citep{jiang2021xlm}, a knowledge-enhanced multilingual LM proposed in a recent work. Even when compared with XLM-R$_\textrm{LARGE}$, our large model still improves the average performance by 1.56. The WikiAnn dataset allows us to evaluate our models on all of the 10 languages involved in pretraining. \citet{jiang2021xlm} did not report XLM-K results on WikiAnn, so we evaluate their pretrained model on WikiAnn and the following knowledge intensive tasks for better comparison. As the results shown in Table~\ref{tab:exp_ner_wikiann}, our best base and large models outperform the corresponding XLM-R models by 2.64 and 1.60 respectively.
From both datasets we observe KMLM$_\textrm{CS}$-XLM-R$_\textrm{BASE}$ performs better than KMLM$_\textrm{Parallel}$-XLM-R$_\textrm{BASE}$, which shows the efficacy of the code-switching technique for large-scale cross-lingual pretraining. 
Moreover, both KMLM$_\textrm{Mix}$-XLM-R$_\textrm{BASE}$ and KMLM$_\textrm{Mix}$-XLM-R$_\textrm{LARGE}$ (i.e. the models pretrained on the mixed code-switched and parallel data) surpass all of the compared models in terms of F1, suggesting that the mixed data can help further generalize the representations across languages.

\subsection{Factual Knowledge Retrieval}
\label{sec:exp_xfactr}

X-FACTR~\citep{jiang-etal-2020-x} is a benchmark for assessing the capability of multilingual pretrained language model on capturing factual knowledge. It provides multilingual cloze-style question templates and the underlying idea is to query knowledge from the models for filling in the blank of these question templates. From \citep{jiang-etal-2020-x}, we notice the performance of XLM-R$_\textrm{BASE}$ is much worse than mBERT$_\textrm{BASE}$ (see Table~\ref{tab:exp_xfactr}). It is probably because mBERT$_\textrm{BASE}$ has a much smaller vocabulary than XLM-R (120k vs 250k) and employs Wikipedia corpus instead of the general data crawled from the Internet. So we also pretrain KMLM$_\textrm{CS}$-mBERT$_\textrm{BASE}$ for more comprehensive comparison. As we can see from Table~\ref{tab:exp_xfactr}, all of the models trained with our framework demonstrate significant improvements on factual knowledge retrieval accuracy, which again indicates the benefits of our method on factual knowledge acquisition. Our model still demonstrates better performance than XLM-K, though it is also trained using Wikipedia.

\begin{table*}[t]
    \centering
    \scalebox{0.66}{
    \begin{tabular}{lccccccccc}
    \Xhline{3\arrayrulewidth}
         & en & es & fr & nl & ja & ko & vi & zh & avg   \\ \hline
        XLM-K & 7.7 & \textbf{7.3} & 3.6 & 5.0 & 0.3 & 4.0 & 5.0 & 0.9 & 4.2 \\
        XLM-R$_\textrm{BASE}$ & 4.5 & 3.1 & 2.0 & 1.6 & \textbf{1.8} & 2.1 & 3.6 & 1.0 & 2.5 \\
        KMLM$_\textrm{CS}$-XLM-R$_\textrm{BASE}$ (ours) & \textbf{8.6} & 4.8 & 4.2 & 5.6 & 1.6 & 4.2 & 5.8 & 3.0 & 4.7 \\
        KMLM$_\textrm{Parallel}$-XLM-R$_\textrm{BASE}$ (ours) & 8.1 & 5.4 & 3.7 & \textbf{6.1} & 1.6 & 4.7 & \textbf{6.3} & 2.4 & 4.9 \\ 
        KMLM$_\textrm{Mix}$-XLM-R$_\textrm{BASE}$ (ours) & 7.9 & 5.1 & \textbf{4.8} & \textbf{6.1} & 1.7 & \textbf{4.8} & 6.2 & \textbf{3.1} & \textbf{5.0} \\ 
        \hline
        XLM-R$_\textrm{LARGE}$ & 7.9 & 4.4 & 3.8 & 5.0 & \textbf{2.9} & 5.2 & 5.7 & 1.0 & 4.5 \\ 
        KMLM$_\textrm{CS}$-XLM-R$_\textrm{LARGE}$ (ours) & 10.5 & 5.5 & 6.9 & 7.1 & 1.1 & 6.7 & 5.7 & 1.6 & 5.6 \\
        KMLM$_\textrm{Mix}$-XLM-R$_\textrm{LARGE}$ (ours) & \textbf{11.1} & \textbf{5.8} & \textbf{7.3} & \textbf{7.7} & 1.4 & \textbf{7.1} & \textbf{6} & \textbf{3.8} & \textbf{6.3} \\
        \hline
        mBERT$_\textrm{BASE}$ & 8.4 & 8.7 & 5.5 & 8.6 & 1.0 & 2.0 & 4.7 & 4.5 & 5.4 \\
        KMLM$_\textrm{CS}$-mBERT$_\textrm{BASE}$ (ours) & \textbf{13.0} & 10.9 & 8.5 & \textbf{11.8} & 2.0 & \textbf{3.2} & \textbf{10.1} & 10.7 & 8.8 \\
        KMLM$_\textrm{Mix}$-mBERT$_\textrm{BASE}$ (ours) & 12.5 & \textbf{11.3} & \textbf{8.7} & 11.7 & \textbf{2.2} & 3 & 9.9 & \textbf{11.6} & \textbf{8.9} \\
    \Xhline{3\arrayrulewidth}
    \end{tabular}}
    \caption{Factual knowledge retrieval results (acc., \%) on X-FACTR.}
    \label{tab:exp_xfactr}
\end{table*}

\subsection{Cross-lingual Relation Classification}

RELX~\citep{koksal-ozgur-2020-relx} is developed by selecting a subset of KBP-37~\citep{zhang2015relation}, a commonly-used English relation classification dataset, and by generating human translations and annotations in French, German, Spanish, and Turkish. We evaluate the same set of models as \S\ref{sec:exp_xfactr}, since mBERT$_\textrm{BASE}$ also outperforms XLM-R$_\textrm{BASE}$ on this task. The evaluation script provided by \citet{koksal-ozgur-2020-relx} is used to finetune the pretrained models on English training set and evaluate on the target language test sets. As the results shown in Table~\ref{tab:exp_relx}, all of our models achieves consistently higher accuracy than XLM-K and XLM-R.

\begin{table}[t!]
    \centering
    \scalebox{0.6}{
    \begin{tabular}{lcccccc}
    \Xhline{3\arrayrulewidth}
        & en & es & de & fr & avg$_{tgt}$ & $\Delta_{avg}$ \\ \hline
        XLM-K & 59.1 & \textbf{58.3} & 55.9 & 56.4 & 57.4 & - \\
        XLM-R$_\textrm{BASE}$ & \textbf{62.7} & 55.1 & 54.8 & 54.3 & 54.7 & 0  \\
        KMLM$_\textrm{CS}$-XLM-R$_\textrm{BASE}$ (ours) & 61.2 & 57.9 & 56.6 & 55.9 & 57.9 & +3.2 \\ 
        KMLM$_\textrm{Parallel}$-XLM-R$_\textrm{BASE}$ (ours) & 62.6 & 56.6 & \textbf{56.9} & 55.0 & 57.8 & +3.1 \\
        KMLM$_\textrm{Mix}$-XLM-R$_\textrm{BASE}$ (ours) & 61.6 & 56.8 & 57 & \textbf{58.4} & \textbf{58.5} & +3.8 \\
        \hline
        XLMR$_\textrm{LARGE}$ & 62.8 & 62.6 & 60.4 & 59.5 & 60.8  & 0 \\
        KMLM$_\textrm{CS}$-XLM-R$_\textrm{LARGE}$ (ours) & 63.5 & \textbf{63.7} & 60.1 & 60.4 & \textbf{61.4} & +0.6 \\
        KMLM$_\textrm{Mix}$-XLM-R$_\textrm{LARGE}$ (ours) & \textbf{63.6} & 61.7 & \textbf{61.8} & \textbf{60.7} & \textbf{61.4} & +0.6 \\
        \hline
        mBERT$_\textrm{BASE}$ & \textbf{65.8} & 58.9 & 58.5 & 58.2 & 58.5 & 0 \\
        KMLM$_\textrm{CS}$-mBERT$_\textrm{BASE}$ (ours) & 64.2 & 59.5 & \textbf{59.1} & \textbf{61.7} & \textbf{60.1} & +1.6 \\ 
        KMLM$_\textrm{Mix}$-mBERT$_\textrm{BASE}$ (ours) & 60.9 & \textbf{61.3} & 57.8 & 60.3 & 59.8 & +1.3 \\
    \Xhline{3\arrayrulewidth}
    \end{tabular}}
    \caption{Zero-shot cross-lingual relation classification performance (acc., \%) on RELX.}
    \label{tab:exp_relx}
\end{table}

\subsection{Cross-lingual Logical Reasoning}
\label{sec:crosslingual_logical_reason}

\begin{figure}[t]
    \centering
    \resizebox{0.98\linewidth}{!}{    
    \adjustbox{minipage=[r][17em][b]{0.8\textwidth},scale={0.9}}{
    \textcolor{blue}{Context:} \\
    (Poland, located in time zone, UTC+01:00)\\
    (Poland, located in time zone, Central European Time)\\
    
    \textcolor{blue}{Question:} \\
    What is the relation between UTC+01:00 and Central European Time? \\
    
    \textcolor{blue}{Choices:} \\
    part of, \textcolor{orange}{said to be the same as}, located in time zone, instance of, has part, followed by\\
    
    \textcolor{blue}{Answer:} \\ 
    \textcolor{orange}{said to be the same as}
    }}
    \caption{An example (English) extracted from our cross-lingual logical reasoning (XLR) dataset.}
    \label{fig:xlr_data}

\end{figure}

\begin{table*}[t!]
    \centering
    \scalebox{0.66}{
    \begin{tabular}{lcccccccccccc}
    \Xhline{3\arrayrulewidth}
        & en & de & es & fr & it & ja & ko & nl & vi & zh & avg$_{tgt}$ & $\Delta_{avg}$ \\ 
        \hline
        XLM-K & 58.29 & 44.19 & 44.41 & 45.14 & 41.02 & 25.94 & 34.32 & 46.92 & 38.19 & 29.81 & 38.88 & - \\
        XLM-R$_\textrm{BASE}$ & 64.38 & 46.38 & 49.36 & 45.46 & 46.38 & 21.58 & 32.53 & 54.06 & 39.11 & 27.80 & 40.30 & 0 \\ 
        KMLM$_\textrm{CS}$-XLM-R$_\textrm{BASE}$ (ours) & 71.52 & 63.52 & 64.73 & \textbf{66.19} & 63.68 & \textbf{37.04} & \textbf{43.39} & 63.42 & 55.17 & \textbf{45.77} & \textbf{55.88} & +15.58 \\ 
        KMLM$_\textrm{Parallel}$-XLM-R$_\textrm{BASE}$ (ours) & 70.03 & 60.48 & 61.78 & 62.38 & 62.95 & 35.21 & 44.76 & 62.03 & \textbf{57.75} & 42.60 & 54.44 & +14.14 \\
        KMLM$_\textrm{Mix}$-XLM-R$_\textrm{BASE}$ (ours) & \textbf{72.70} & \textbf{63.71} & \textbf{66.41} & 65.05 & \textbf{64.98} & 36.16 & 38.79 & \textbf{64.83} & 56.51 & 44.73 & 55.69 & +15.39 \\
        \hline
        XLM-R$_\textrm{LARGE}$ & 79.39 & 68.38 & 73.46 & 71.20 & 70.82 & 56.25 & 47.61 & 70.98 & 66.00 & 55.11 & 64.42 & 0 \\ 
        KMLM$_\textrm{CS}$-XLM-R$_\textrm{LARGE}$ (ours) & \textbf{87.07} & \textbf{85.87} & 83.58 & \textbf{86.03} & \textbf{83.80} & 75.04 & 75.39 & 85.01 & \textbf{83.58} & \textbf{83.74} & \textbf{82.45} & +18.03 \\ 
        KMLM$_\textrm{Mix}$-XLM-R$_\textrm{LARGE}$ (ours) & 86.67 & 84.25 & \textbf{83.75} & 85.14 & 82.89 & \textbf{76.03} & \textbf{77.30} & \textbf{85.30} & 82.86 & 82.57 & 82.23 & +17.81 \\ 
    \Xhline{3\arrayrulewidth}
    \end{tabular}
    }
    \caption{Zero-shot cross-lingual logical reasoning performance (acc., \%).}
    \label{tab:exp_logic}
\end{table*}

\paragraph{Dataset}
To verify the effectiveness of our logical reasoning oriented pretraining tasks (\S\ref{sec:pretrain_task}) in an intrinsic way, we propose a cross-lingual logical reasoning (XLR) task in the form of multiple-choice questions. An example of such reasoning question is given in Fig.~\ref{fig:xlr_data}. 
The dataset is constructed using the cycles extracted from Wikidata. We manually annotate 1,050 samples in English and then translated them to the other 9 non-English languages (see Sec. \ref{subsec:pre}) to build the multilingual test sets. The 3k train samples and 1k dev samples in English are also generated and cleaned automatically. The cycles used to build the test set are removed from the pretraining data, so our PLMs have never seen them beforehand. The detailed dataset construction steps can be found in Appendix~\S\ref{sec:app_xlr}. 

\begin{figure}[t!]
\centering
\vspace{-1em}
\includegraphics[width=0.47\columnwidth]
{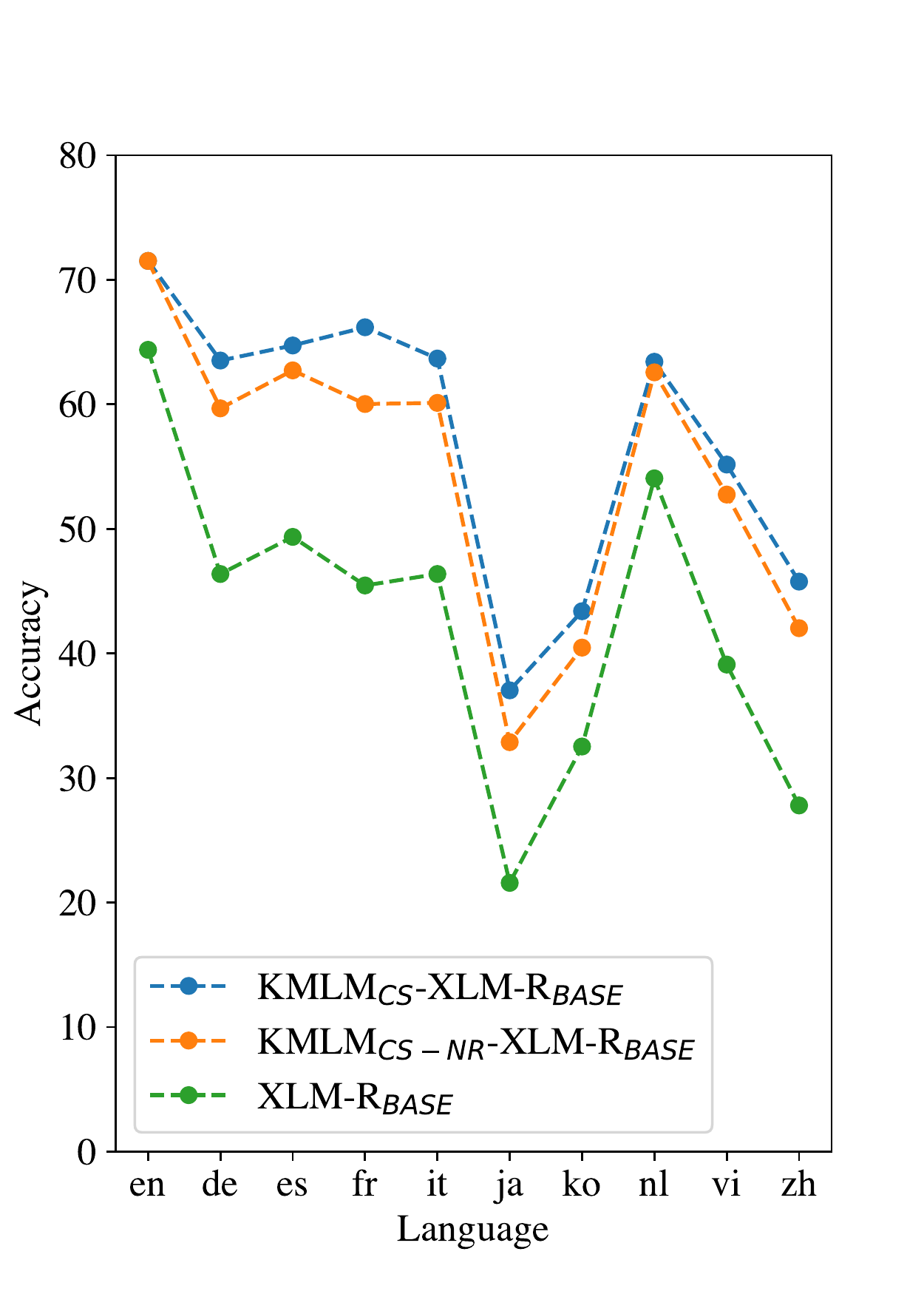}
\hspace{-0.5em}
\includegraphics[width=0.47\columnwidth]{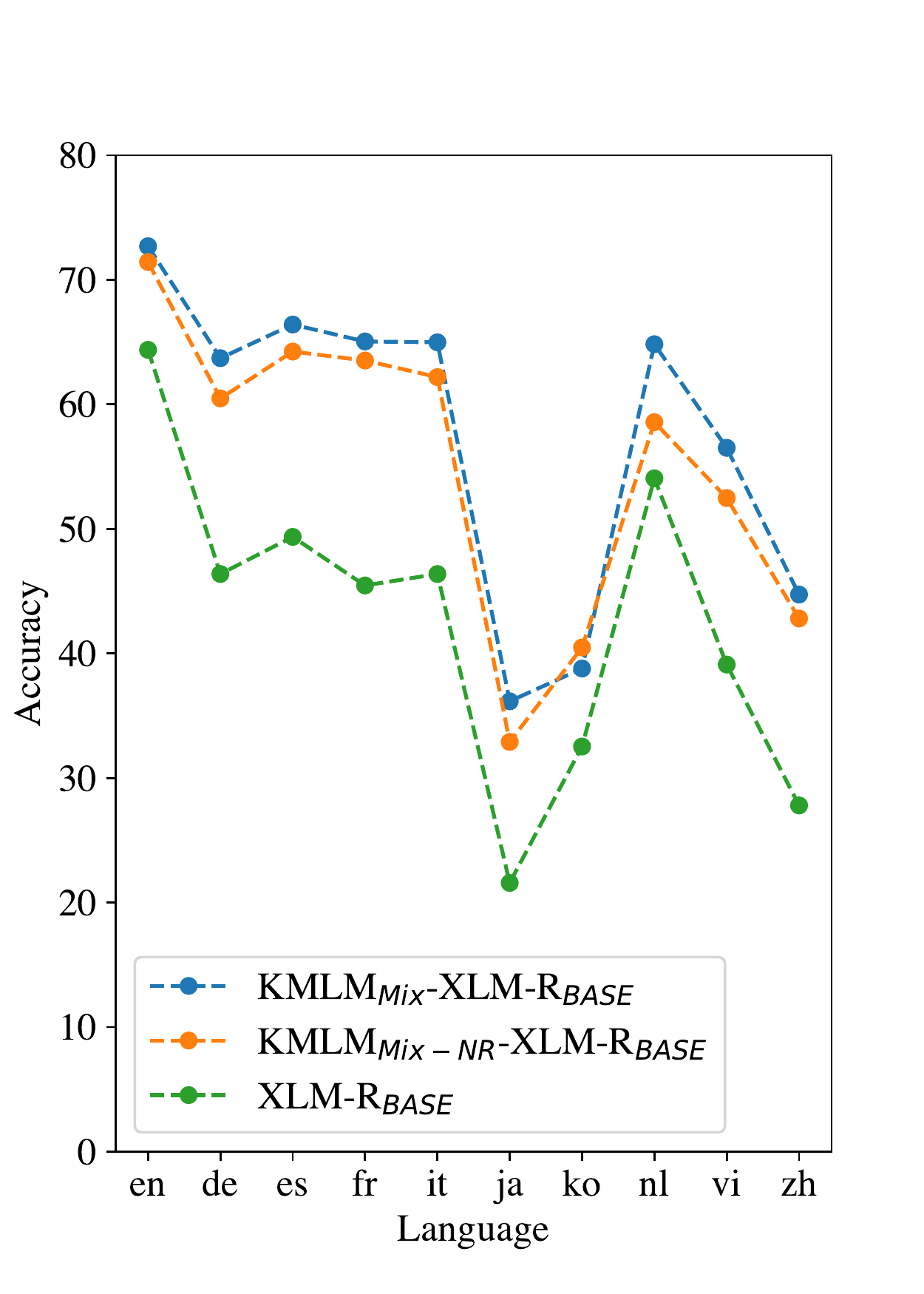}
\caption{Comparison of the models trained with and without logical reasoning task.}
\label{fig:ablation_logic}
\end{figure}

\paragraph{Results}
We modify the multiple choice evaluation script implemented by Hugging Face\footnote{\url{https://github.com/huggingface/transformers/tree/master/examples/pytorch/multiple-choice}} for this experiment. The models are finetuned on the English training set, and evaluated on the test sets in different target languages. Results are presented in Table~\ref{tab:exp_logic}. All of our models outperform the baselines significantly. Unlike on the previous tasks, where KMLM$_\textrm{Mix}$ often performs the best, KMLM$_\textrm{CS}$ shows slightly higher accuracy than KMLM$_\textrm{Mix}$.
We also conduct ablation study to verify the effectiveness of our proposed logical reasoning oriented pretraining task. We pretrain the \textbf{N}one-\textbf{R}easoning models, KMLM$_\textrm{CS-NR}$-XLM-R$_\textrm{BASE}$ and KMLM$_\textrm{Mix-NR}$-XLM-R$_\textrm{BASE}$ on the same data as KMLM$_\textrm{CS}$-XLM-R$_\textrm{BASE}$ and KMLM$_\textrm{Mix}$-XLM-R$_\textrm{BASE}$, but without the logical reasoning tasks, i.e., with the MLM task only on the reasoning-based data. As presented in Fig.~\ref{fig:ablation_logic}, the none-reasoning models also performs better than XLM-R$_\textrm{BASE}$, which shows the usefulness of our reasoning-based data. We also observe KMLM$_\textrm{CS}$-XLM-R$_\textrm{BASE}$ and KMLM$_\textrm{Mix}$-XLM-R$_\textrm{BASE}$, i.e., the models pretrained with logical reasoning tasks, consistently perform the best, which proves our proposed task can help models learn logical patterns more efficiently.

\subsection{General Cross-lingual Tasks}
\label{sec:general_crosslingual}

\begin{table}[t]
    \centering
    \scalebox{0.5}{
    \begin{tabular}{lcccccc}
    \Xhline{3\arrayrulewidth}

     & POS & XQuAD & MLQA & TyDiQA & XNLI & PAWSX \\ %\hline
     Metrics & F1 & F1/EM & F1/EM & F1/EM & Acc. & Acc.\\ \hline
     XLM-R$_\textrm{BASE}$ & \textbf{72.8} & \textbf{69.6}/\textbf{53.9} & 64.9/47.2 & 45.2/28.5 & \textbf{73.7} & 84.8 \\
     KMLM$_\textrm{CS}$-XLM-R$_\textrm{BASE}$ (ours) & 71.2 & 69.2/53.3 & 64.7/47.0 & 48.7/29.2 & 73.6 & \textbf{85.1} \\ 
     KMLM$_\textrm{Parallel}$-XLM-R$_\textrm{BASE}$ (ours) & 71.2 & 69.3/53.4 & 64.7/46.8 & \textbf{51.4}/\textbf{32.8} & 73.3 & 84.1 \\
     KMLM$_\textrm{Mix}$-XLM-R$_\textrm{BASE}$ (ours) & 71.4 & 69.5/53.3 & \textbf{65.4}/\textbf{47.3} & 49.6/32.6 & 72.9 & 84.7 \\ 
     \hline
     MMTE$^\dag$ & 73.5 &  64.4/46.2 & 60.3/41.4 &  58.1/43.8 & 67.4 & 81.3 \\
     mBERT$_\textrm{LARGE}^\dag$ & 70.3 & 64.5/49.4 & 61.4/44.2 & 59.7/43.0 & 65.4 & 81.9 \\
     XLM-R$_\textrm{LARGE}$ & \textbf{74.6} & 76.8/60.9 & \textbf{72.5}/\textbf{54.2} & 66.6/46.6 & 79.0 & 87.8 \\
     KMLM$_\textrm{CS}$-XLM-R$_\textrm{LARGE}$ (ours) & 72.4 & 76.5/60.6 & 72.0/53.7 & 66.4/47.9 & 78.6 & 87.7 \\
     KMLM$_\textrm{Mix}$-XLM-R$_\textrm{LARGE}$ (ours) & 72.8 & \textbf{77.3}/\textbf{61.7} & 72.1/53.7 & \textbf{67.9}/\textbf{50.4} & \textbf{79.2} & \textbf{88.0} \\
    \Xhline{3\arrayrulewidth} 
    \end{tabular}}
    \caption{Zero-shot cross-lingual POS, QA and classification results. Note that the performance of the languages not appearing in our prepared pretraining data are also counted.  $^\dag$The results are from \citep{hu2020xtreme}.}
    \label{tab:exp_xtreme}
\end{table}

Recall that our models are directly trained on the structured KG data. Though we attempt to minimize its difference from the natural sentences when designing the pretraining tasks, it is unknown how the difference affects cross-lingual transfer performance on the general NLP tasks. Therefore, we also evaluate our models on the part-of-speech (POS) tagging, question answering and classification tasks prepared by XTREME~\citep{hu2020xtreme}. Experimental results are shown in Table~\ref{tab:exp_xtreme}. Note that many of the languages covered by these tasks are not in our pretraining data, but we include all their results when computing the average performance. Overall, the performance of our models is comparable with the baselines on all of the tasks, except POS. Possibly because the POS task is more sensitive to the change of the training sentence structures. Though some of our models perform slightly better than the baselines on XQuAD \citep{Artetxe2019xquad} and MLQA \citep{lewis2019mlqa}, we find the performance gain of our models on TyDiQA\footnote{Same as XTREME, we use the gold
passage version of TyDiQA.} \citep{clark2020tydi} is more obvious, which is a more challenging QA task that has less lexical overlap between question-answer pairs. From these results we can see that, when our KMLMs achieve consistent improvements on the knowledge-intensive tasks, as shown by the experimental results in the previous subsections, it does not sacrifice the performance on the general NLP tasks.

\section{Conclusions}

In this paper, we have presented a novel framework for knowledge-based multilingual language pretraining. 
Our approach firstly creates a synthetic multilingual corpus from the existing KG and then tailor-makes two pretraining tasks, multilingual knowledge oriented pretraining and logical reasoning oriented pretraining. These multilingual pretraining tasks not only facilitate factual knowledge memorization but also boost the capability of implicit knowledge modeling. 
We evaluate the proposed framework on a series of knowledge-intensive cross-lingual tasks and the comparison results consistently demonstrate its effectiveness. 

\section*{Limitations}
The KMLM models proposed in this work are pretrained on 10 languages in our experiments, so it is unclear whether scaling up to more languages will help further improve its performance on the downstream tasks. Due to the high computation cost, we leave it for future work. Despite the promising performance improvement on the knowledge intensive tasks, we also observe that KMLM do not perform well on the part-of-speech tagging tasks (\S\ref{sec:general_crosslingual}). It is possibly caused by the large amount of synthetic sentences used in pretraining, where mask tokens are used to replace the linking words. In future, we will  explore efficient ways to leverage pretrained denoising models \citep{liu-etal-2020-multilingual-denoising} or graph-to-sequence models \citep{NEURIPS2021_1e747ddb} to convert the synthetic sentences or knowledge triples to the form more close to natural sentences.

\section*{Ethical Impact}

Neural models have achieved significant success in many NLP tasks, especially for the popular languages like English, Spanish, etc. However, neural models are data hungry, which poses challenges for applying them to the low-resource languages due to the limited NLP resources. In this work, we propose methods to inject knowledge into the multilingual pretrained language models, and enhance their logical reasoning ability. Through extensive experiments, our methods have been proven effective in a wide range of knowledge intensive multilingual NLP tasks. Therefore, our proposed method could help overcome the resource barrier, and enable the advances in NLP to benefit a wider range of population.

\section*{Acknowledgements}
\label{sec:acknowledgements}

This research is partly supported by the Alibaba-NTU Singapore Joint Research Institute, Nanyang Technological University. Linlin Liu would like to thank the support from Interdisciplinary Graduate School, Nanyang Technological University.

% Entries for the entire Anthology, followed by custom entries
\bibliography{main}
\bibliographystyle{acl_natbib}

\appendix
\section{Appendix}

\subsection{Language Model Pretraining Details}
\label{sec:app_pretrain}

\paragraph{Training Data}
The statistics of the data used for pretraining are shown in Table~\ref{tb:app_pretrain_data}.

\begin{table}[htbp]
\vspace{0.4em}
\centering
\scalebox{0.7}{\begin{tabular}{lc}
\toprule
\bf{Description} & \bf{Number} \\
\midrule
languages & 10 \\
code switched synthetic sentences & 246,783,693 \\
parallel synthetic sentences & 190,576,098 \\
unique relation combinations in length-3 cycles & 29,819 \\
unique relation combinations in length-4 cycles & 239,966 \\
reasoning based training samples from length-3 cycles & 24,142,272 \\
reasoning based training samples from length-4 cycles & 73,881,422 \\
sampled CC100 sentences (KMLM-XLM-R only) & 260,000,000 \\
sampled Wikipedia sentences (KMLM-mBERT only) & 153,011,930 \\
\bottomrule
\end{tabular}}
\caption{Statistics of the data used for pretraining.}
\label{tb:app_pretrain_data}
\end{table}

\paragraph{Hyper-Parameters}
The hyper-parameters used for language model pretraining are presented in Table~\ref{tb:app_hyperparameter}. After pretraining, we finetune the models on the plain text data with max sequence length of 512 for another 600 steps. Due to the high computation cost of LM pretraining, we do not run many experiments for hyper-parameter searching. Instead, the learning rate, batch size, mlm probability are determined according to those used in the previous LM pretraining studies. To determine the knowledge task loss weight $\alpha$ for large scale pretraining, we compare $\alpha \in \{0.5, 0.3, 0.1\}$ using the base models pretrained on a smaller dataset. Each base model takes about 30 days to train with 8 V100 GPUs.

\begin{table}[htbp]
\vspace{0.5em}
\centering
\scalebox{0.8}{\begin{tabular}{lc}
\toprule
\bf{Hyper-parameter} & \bf{Value} \\
\midrule
learning rate & 5e-5 \\
weight decay & 0 \\
optimizer & AdamW \\
number of train epochs & 1 \\
batch size for the natural sentences & 9,600 \\
batch size for code switched knowledge data & 9,600 \\
batch size for reasoning data & 9,600 \\
mlm probability & 0.15 \\
max sequence length & 128 \\
number of warmup steps & 100 \\
knowledge task loss weight ($\alpha$ in the loss function) & 0.3 \\
\bottomrule
\end{tabular}}
\caption{Hyper-parameters used for language model pretraining.}
\label{tb:app_hyperparameter}
\end{table}

\subsection{Cross-lingual Logical Reasoning Task}
\label{sec:app_xlr}

We propose a cross-lingual logical reasoning (XLR) task in the form of multiple-choice questions to verify the effectiveness of our logical reasoning oriented pretraining tasks in an intrinsic way. An example of such reasoning question is given in Fig.~\ref{fig:xlr_data}. 
The dataset is constructed using the length-3 and length-4 cycles extracted from Wikidata. 
For each cycle, we pick a triplet to create the question and answer. The question is created by asking the relation between a pair of entities in that triplet. 6 choices are provided for each question (including the correct answer), which contains all of the relations appear in the cycle and some sampled relations associated with the two entities. The remaining triplets from the cycle are used as the context, which is in the form of knowledge graph (see Fig.~\ref{fig:xlr_data}).
The model is required to select the most probable choice according to the given context and question.
We provide correct and incorrect examples to the annotators, and manually annotate 1,050 samples in English to build the test set. The train and dev sets are automatically generated, and then cleaned by balancing the appearances of entities, relations and answers. After cleaning, we randomly select 3k train samples and 1k dev samples for the experiment. Then the multilingual test data in the other 9 non-English languages are generated by selecting the entity/relation labels in the desired languages from Wikidata. The cycles used to build the test set are removed from the pretraining data, so our PLMs have never seen them beforehand.

Statistics of the self-constructed cross-lingual logic reasoning (XLR) dataset are presented in Table~\ref{tb:app_reasoning_data}. The multilingual test data in the 9 non-English languages are generated by selecting the entity/relation labels in the desired languages from Wikidata. So the statistics for their test sets are the same as English.

\begin{table}[h]
\vspace{0.5em}
\centering
\scalebox{1}{\begin{tabular}{lc}
\toprule
\bf{Description} & \bf{Value} \\
\midrule
number of samples in the train set  & 3,000 \\
number of samples in the dev set  & 1,000 \\
number of samples in the test set  & 1,050 \\
train set unique relation combinations & 1,419 \\
dev set unique relation combinations & 746 \\
test set unique relation combinations & 444 \\
\bottomrule
\end{tabular}}
\caption{Statistics of the self-constructed cross-lingual logic reasoning data (English).}
\label{tb:app_reasoning_data}
\end{table}

\subsection{Impact of the Logical Reasoning Tasks}
\label{sec:app_xlr_impact}

As discussed in \S\ref{sec:crosslingual_logical_reason}, we pretrain the \textbf{N}one-\textbf{R}easoning models, KMLM$_\textrm{CS-NR}$-XLM-R$_\textrm{BASE}$ and KMLM$_\textrm{Mix-NR}$-XLM-R$_\textrm{BASE}$ on the same data as KMLM$_\textrm{CS}$-XLM-R$_\textrm{BASE}$ and KMLM$_\textrm{Mix}$-XLM-R$_\textrm{BASE}$, but without the logical reasoning tasks. The none-reasoning models generally perform worse than the corresponding models trained with the logical reasoning tasks, which proves the usefulness of the tailored logical reasoning oriented pretraining task for logical reasoning.

In order to explore the impact of the logical reasoning oriented pretraining tasks on the none-logical reasoning tasks, we also conduct ablation studies to compare the performance of KMLM$_\textrm{CS-NR}$-XLM-R$_\textrm{BASE}$ and KMLM$_\textrm{CS}$-XLM-R$_\textrm{BASE}$ on the CoNLL02/03 \citep{tjong-kim-sang-2002-introduction,tjong-kim-sang-de-meulder-2003-introduction} and WikiAnn \citep{pan-etal-2017-cross} NER data. From the results presented in Table~\ref{tab:app_lr_ner_conll} and \ref{tab:app_lr_ner_wikiann} we can see that the average performance on the target languages are very close, which shows the logical reasoning oriented pretraining tasks do not have obvious impact on zero-shot cross-lingual NER.

\begin{table}[t!]
    \centering
    \scalebox{0.66}{
    \begin{tabular}{lcccccc}
    \Xhline{3\arrayrulewidth}
        & en & de & nl & es & avg$_{tgt}$ \\ 
        \hline
        KMLM$_\textrm{CS}$-XLM-R$_\textrm{BASE}$ & 91.47 & 73.52 & 80.95 & 76.59 & 77.02  \\
        KMLM$_\textrm{CS-NR}$-XLM-R$_\textrm{BASE}$ & 91.38 & 73.76 & 81.55 & 76.03 & 77.11 \\
    \Xhline{3\arrayrulewidth}
    \end{tabular}}
    \caption{Zero-shot cross-lingual NER F1 on the CoNLL02/03 datasets.}
    \label{tab:app_lr_ner_conll}
\end{table}

\begin{table}[t!]
    \centering
    \scalebox{0.62}{
    \begin{tabular}{lcccccc}
    \Xhline{3\arrayrulewidth}
        & en & vi & nl & de & fr & it \\ 
        \hline
        KMLM$_\textrm{CS}$-XLM-R$_\textrm{BASE}$ & 83.43 & 70.55 & 82.18 & 77.87 & 79.19 & 80.06 \\
        KMLM$_\textrm{CS-NR}$-XLM-R$_\textrm{BASE}$ & 83.75 & 70.73 & 82.44 & 78.03 & 78.88 & 80.20 \\
        \hline
        \hline
        & es & ja & ko & zh & avg$_{tgt}$ \\
        \hline
        KMLM$_\textrm{CS}$-XLM-R$_\textrm{BASE}$ & 75.96 & 19.32 & 57.54 & 20.95  & 62.62 & \\
        KMLM$_\textrm{CS-NR}$-XLM-R$_\textrm{BASE}$ & 74.97 & 18.39 & 58.16 & 20.58  & 62.49 & \\
    \Xhline{3\arrayrulewidth}
    \end{tabular}
    }
    \caption{Zero-shot cross-lingual NER F1 on the WikiAnn dataset.}
    \label{tab:app_lr_ner_wikiann}
\end{table}

\end{document}